\documentclass{article}
\usepackage{spconf,amsmath,graphicx,hyperref}
\usepackage{enumitem}
\usepackage{amssymb, amsfonts, bm}
\usepackage{algorithm}
\usepackage{siunitx}
\usepackage{algpseudocode, booktabs}
\usepackage[T1]{fontenc}

\usepackage{tabularx}
\usepackage[numbers]{natbib}
\usepackage{adjustbox}

\title{Learning Longitudinal Health Representations from EHR and Wearable Data}
\name{Yuanyun Zhang, Han Zhou, Li Feng, Yilin Hong, Shi Li}
\address{University of the Chinese Academy of Sciences, Columbia University }
\begin{document}
\onecolumn
%
\maketitle
\begin{abstract}

Foundation models trained on electronic health records (EHRs) have shown strong performance across a wide range of clinical prediction tasks, yet they remain fundamentally limited by the sparse, irregular, and event-driven nature of clinical documentation. In parallel, wearable devices generate dense, continuous physiological time series that capture subclinical dynamics but lack semantic grounding and task-specific supervision. Existing approaches typically model these modalities in isolation or combine them through late fusion, failing to exploit their complementary statistical structure. In this work, we introduce a multimodal foundation model that jointly represents EHR data and wearable sensor streams as a unified continuous-time latent process. Our approach combines modality-specific encoders with a shared temporal backbone and is pretrained using self-supervised and cross-modal objectives that align dense physiological signals with sparse clinical events. By explicitly modeling time and enforcing asymmetric cross-modal prediction, the model learns representations that are both temporally coherent and clinically grounded. We evaluate the learned representations across clinical event forecasting, physiological state estimation, and longitudinal risk modeling tasks, using frozen probes to isolate pretraining effects. Across all settings, the proposed multimodal model consistently outperforms strong EHR-only and wearable-only foundation models, including MOTOR, CEHR-BERT, PaPaGei, and HiMAE, with particularly large gains at long prediction horizons and under realistic missing-modality scenarios. Representation analyses further show smoother latent trajectories and improved calibration relative to unimodal baselines. Our results demonstrate that joint EHR–wearable pretraining yields a qualitatively different class of patient representations, bridging sparse clinical observability and continuous physiological dynamics. This work suggests that multimodal foundation models are a necessary step toward faithful, transferable representations of longitudinal health.

\end{abstract}

\section{Introduction}
\label{sec:intro}

The past year has seen the rapid emergence of foundation models \citep{he2024foundation, awais2025foundation, thakur2024foundation} as a unifying paradigm for representation learning across natural language, vision, and increasingly, biomedical data. In healthcare, this shift has been most visible in the success of large-scale models trained on electronic health records (EHRs) \citep{evans2016electronic, hoerbst2010electronic, seymour2012electronic}, which leverage the longitudinal, high-dimensional, and semantically rich structure of clinical data to learn reusable representations for downstream prediction tasks. In parallel, the proliferation of consumer and clinical-grade wearable devices has produced a complementary stream of high-frequency physiological measurements, capturing aspects of human health that are largely invisible to traditional clinical documentation \citep{soumma2024wearable, narayanswamyscaling}. Despite their shared goal of characterizing patient state over time, EHRs and wearables have largely been modeled in isolation. This separation reflects not only historical data silos, but also deeper methodological challenges arising from mismatched temporal resolutions, observation mechanisms, and inductive biases.

At a conceptual level, EHR data and wearable data encode distinct but interdependent views of health. EHRs provide sparse, event-driven observations mediated by clinical workflows, such as diagnoses, laboratory tests, medications, and procedures. These observations are irregularly sampled, often missing not at random, and semantically structured through medical ontologies and billing codes. Wearable data, by contrast, offers dense, continuous-time measurements of physiological signals such as heart rate, activity, sleep, and autonomic dynamics. These signals reflect underlying biological processes at timescales ranging from seconds to days, often preceding clinical manifestation by weeks or months. From an information-theoretic perspective, the two modalities occupy different regions of the signal-to-noise spectrum: EHRs are low-frequency but high-semantic-content measurements, while wearables are high-frequency but weakly labeled proxies of latent health state. Any model that seeks to represent patient trajectories faithfully must therefore reconcile these complementary regimes.

The central thesis of this work is that jointly modeling EHR and wearable data within a unified foundation model framework enables the learning of richer, more temporally grounded patient representations than either modality alone. The motivation is not merely additive performance gains on downstream tasks, but a qualitative shift in what aspects of health can be represented. Many clinically meaningful phenomena, such as disease onset, exacerbation, recovery, and treatment response, unfold continuously over time but are only intermittently recorded in the EHR. Wearable signals can provide dense contextualization of these latent transitions, while EHR events anchor them to clinically interpretable states. In this sense, the modalities are not redundant but conditionally informative: wearable data disambiguates the temporal dynamics between EHR events, while EHR data provides semantic supervision that stabilizes and grounds wearable representations.

However, integrating these modalities poses nontrivial challenges that preclude straightforward concatenation or late fusion. The data-generating processes differ fundamentally. Let \(x_i(t)\) denote the latent health state of patient (i) at time (t). EHR observations \(y_i^\text{EHR}(t_k)\) occur at discrete, irregular times \(t_k\) and are conditionally dependent on both \(x_i(t_k)\) and clinical decision processes, while wearable observations \(y_i^\text{wear}(t)\) form a dense stochastic process with different noise characteristics and partial observability. Any joint model must therefore contend with asynchronous sampling, modality-specific missingness, and scale disparities spanning several orders of magnitude in time. Moreover, supervision is weak and heterogeneous: EHR labels are coarse, delayed, and task-dependent, while wearable data is largely unlabeled. These properties challenge conventional supervised learning formulations and instead point toward self-supervised or weakly supervised pretraining objectives.

Foundation models offer a natural abstraction for addressing these challenges. By pretraining a large-capacity model on multimodal longitudinal data using self-supervised objectives, one can decouple representation learning from task-specific supervision and enable transfer across a wide range of clinical and physiological tasks. Crucially, a multimodal foundation model can exploit cross-modal prediction as an inductive bias, using one modality to predict or contextualize the other. For example, wearable-derived representations may be trained to predict future EHR events, while EHR context may condition the interpretation of physiological dynamics. This cross-modal alignment encourages the model to learn latent factors that are stable across modalities and temporally coherent, approximating the underlying health state trajectory.

The problem we address in this paper is therefore not simply how to combine EHR and wearable features, but how to design a scalable, temporally aware foundation model that respects the statistical structure of each modality while enabling meaningful interaction between them. We argue that success in this setting requires architectural choices that explicitly model time, uncertainty, and modality-specific inductive biases, as well as pretraining objectives that reflect the causal and temporal asymmetries inherent in healthcare data. By framing EHR–wearable integration as a representation learning problem rather than a task-specific engineering exercise, we aim to establish a principled foundation for multimodal patient modeling.

In summary, this work is motivated by a gap between the richness of available longitudinal health data and the representational capacity of current models. EHR-only foundation models risk learning abstractions that are temporally underdetermined, while wearable-only models lack semantic grounding and clinical interpretability. A unified foundation model trained jointly on both modalities has the potential to bridge this gap, yielding representations that are both physiologically sensitive and clinically meaningful. The remainder of this paper develops this argument formally, introduces a multimodal foundation model architecture tailored to EHR and wearable data, and demonstrates its empirical advantages across a range of downstream tasks.

\section{Related works}
\subsection{EHR Foundation Models}

Early EHR foundation models emerged from a straightforward but influential idea: treating structured clinical records as sequences amenable to Transformer-based representation learning \citep{zhang2025chronoformer, tanaka2025temporal, chou2025serialized, huiliang2025clio, ran2025structured, zhang2025collection}. Initial efforts recast longitudinal patient data into token streams, where diagnoses, procedures, and visits were embedded analogously to words in natural language. BEHRT exemplified this paradigm by encoding diagnosis and procedure codes alongside demographic and temporal signals, modeling a patient’s clinical history as a sentence-like object and pretraining with a masked prediction objective on large-scale UK Biobank data \citep{li2020behrt}. This formulation demonstrated that attention-based architectures could recover meaningful temporal dependencies in sparse, irregular medical sequences. Building on this foundation, Med-BERT scaled pretraining to tens of millions of patients and augmented masked modeling with auxiliary supervision related to hospital utilization, while simplifying the embedding design to reduce reliance on explicit temporal or segment encodings \citep{rasmy2021med}.

In parallel with bidirectional masking approaches, several works explored autoregressive formulations better aligned with the inherently sequential nature of care delivery. CEHR-GPT adopted next-event prediction as a primary objective, framing clinical modeling as a generative forecasting problem over coded events \citep{pang2024cehr}. Closely related work retained masked modeling but introduced architectural modifications to better represent time, such as discretized temporal gap tokens and alternative visit-level supervision in place of sentence prediction, as in CEHR-BERT \citep{pmlr-v158-pang21a}. Across these studies, a consistent empirical conclusion emerged: Transformer-based pretraining, whether bidirectional or autoregressive, substantially outperformed recurrent baselines in capturing both semantic co-occurrence and long-range temporal structure in EHR data.

As the field matured, attention shifted from purely representational fidelity toward modeling clinically meaningful dynamics and outcomes. CLMBR introduced an autoregressive framework grounded in OMOP-standardized vocabularies, initially implemented with recurrent architectures and later scaled to a Transformer with over one hundred million parameters \citep{steinberg2021language}. This model demonstrated strong robustness across diverse prediction tasks, including mortality and readmission, and retained performance under temporal and institutional shift. MOTOR extended this trajectory by explicitly parameterizing time-to-event processes, integrating a piecewise exponential survival head into a large Transformer backbone \citep{steinberg2024motor}. By jointly modeling event occurrence and timing across massive cohorts, MOTOR achieved notable improvements in concordance and sample efficiency on datasets ranging from single-institution EHRs to national claims, underscoring the value of explicitly incorporating temporal risk structure into pretraining.

More recent work has focused on mitigating fragmentation arising from heterogeneous coding systems and institutional schemas. Rather than aligning disparate ontologies through manual mappings, text-centric approaches reinterpret structured EHR elements as natural language descriptions. GenHPF adopts this strategy by converting codes, feature names, and numerical values into a hierarchical textual representation processed by a Transformer with event-level temporal aggregation. Related schema-agnostic formulations have appeared in both tabular and multimodal settings \citep{hegselmann2023tabllm, lee2025meme, lee2024emergency}. By operating over semantic text rather than fixed code systems, GenHPF enables unified pretraining across datasets such as MIMIC-III, MIMIC-IV, and eICU without bespoke ontology alignment \citep{johnson2016mimic, johnson2023mimic, pollard2018eicu}. When combined with contrastive and clustering-based objectives, this representation yields strong generalization across tasks and institutions, highlighting semantic normalization as a promising route toward scalable, transferable clinical foundation models \citep{ guo2024multi, ono2024text, an2025dk}

\subsection{Wearable Foundation Models}

Wearable sensing platforms based on photoplethysmography (PPG), electrocardiography (ECG), and inertial measurements continuously capture rich, multivariate time series reflecting cardiovascular function~\citep{castaneda2018review}, physical activity and mobility~\citep{yuan2024self, xu2025lsm}, sleep and circadian rhythms~\citep{li2021transfer, thapa2024sleepfm, logacjov2025long}, and other latent physiological processes. Unlike clinical data, these signals are collected at scale in free-living environments and are almost entirely unlabeled, creating a natural setting for large-scale self-supervised learning~\citep{kaplan2020scaling, bommasani2021opportunities, zhou2024comprehensive, liang2024foundation}. As a result, representation learning for wearable data has increasingly converged on foundation-model-style pretraining regimes.

Self-supervised learning (SSL) dominates this space due to the prohibitive cost and subjectivity of manual annotation. Among SSL objectives, masked reconstruction has become a de facto standard for physiological time series, drawing inspiration from masked language modeling~\citep{devlin-etal-2019-bert} and masked autoencoders in vision~\citep{he2022masked, vaid2023foundational}. In this framework, contiguous segments or patches of the signal are occluded and reconstructed, forcing the model to encode temporal context, periodic structure, and inter-channel dependencies~\citep{zhang2022mask, kong2023understanding, thukral2025wavelet}. Large-scale wearable foundation models, including Google’s LSM family~\citep{narayanswamy2024scaling, xu2025lsm}, rely heavily on this objective across heterogeneous sensor modalities. These masked based objectives models have additionally been seen in deployed scenarios \citep{lee2025towards}.

A complementary line of work employs contrastive learning to impose invariances in representation space~\citep{schmitt2008measurement, jaiswal2020survey}. Here, the central difficulty lies in defining semantically meaningful positive and negative pairs in the absence of labels. One widely adopted heuristic is participant-level contrastive learning, which treats samples from the same individual as positives and samples from different individuals as negatives. This approach underpins several large-scale ECG and PPG foundation models developed in industry settings~\citep{abbaspourazad2023large} and is conceptually aligned with instance discrimination frameworks such as SimCLR~\citep{chen2020simpleframeworkcontrastivelearning}. Other methods inject domain knowledge to define contrastive structure: PaPaGei exploits morphological invariants in PPG waveforms~\citep{pillai2024papagei}, while SleepFM extends contrastive alignment across multiple biosignal modalities, including EEG, ECG, and EMG, to enforce cross-signal consistency~\citep{thapa2024sleepfm}. Additional regularization strategies, such as entropy-based constraints to prevent representation collapse~\citep{jing2021dimcollapse, abbaspourazad2023large}, are often layered on top of these objectives.

Despite their success, contrastive approaches introduce several practical and conceptual limitations. Performance is highly sensitive to the choice of augmentations, many of which lack clear physiological interpretation. Training is computationally demanding due to the need for large batch sizes or memory banks, and the resulting representations are often difficult to interpret, as contrastive objectives provide limited insight into which temporal structures or physiological features are preserved. These limitations have motivated recent interest in hybrid objectives and architectures that combine reconstruction-based learning with temporal prediction or hierarchical modeling, though a consensus on best practices has yet to emerge.

Taken together, prior work on wearable foundation models has established that large-scale SSL can extract meaningful physiological representations from raw sensor streams. However, most existing approaches operate in isolation from clinical context, learning abstractions that are physiologically rich but semantically underconstrained. This separation motivates the multimodal formulations explored in this work, where wearable-derived temporal structure is explicitly grounded in sparse but semantically informative clinical signals.

\subsection{Health FMs in Other Domains}

Beyond electronic health records, the foundation model paradigm has rapidly expanded to encompass a broad spectrum of healthcare data modalities~\citep{ mortazavi2023conference, flores2022conference}, reflecting a field-wide movement toward unified representation learning in medicine~\citep{narayanswamy2024scaling, lin2025case, an2025raptor}. Rather than treating each modality as an isolated prediction problem, these efforts seek to learn reusable latent spaces that generalize across tasks, institutions, and data-generating processes.

In medical imaging, large-scale pretraining on diverse radiographs, pathology slides, and volumetric scans has become standard practice. Models built on convolutional and vision-transformer backbones~\citep{zhang2024challenges, chen2024medical, wang2023real} are trained on millions of images spanning heterogeneous acquisition protocols, scanners, and populations. The resulting representations have demonstrated strong transferability for downstream tasks including disease detection, anatomical segmentation, and outcome prediction, often outperforming task-specific models trained from scratch and exhibiting improved robustness under domain shift.

A parallel line of work has adapted language-model-style objectives to biological sequences. In genomics and molecular biology, DNA, RNA, and protein sequences are treated as symbolic strings, enabling large Transformer models to learn contextual embeddings that encode regulatory structure, evolutionary constraints, and functional motifs~\citep{wu2025generator, lin2025genos, long2025mutbert, guo2025foundation, dalla2025nucleotide}. These models support zero-shot or few-shot inference for tasks such as variant effect prediction, functional annotation, and structure-aware representation learning, illustrating that the foundation model paradigm extends naturally beyond clinical data into molecular-scale biology.

More recently, multimodal foundation models have begun to explicitly integrate heterogeneous healthcare data sources into shared latent representations. Efforts in this direction combine imaging, clinical text~\citep{lee2020biobert, lee2025modern}, physiological waveforms~\citep{abbaspourazad2023large, qiu2025towards, ruan2025ai, abbaspourazad2024wearable}, and structured records within unified architectures designed to support cross-modal alignment and reasoning. By exploiting complementary information across modalities, these models aim to reduce annotation requirements, improve generalization, and enable inference in settings where one or more data sources may be missing or noisy.

Taken together, these developments indicate that foundation modeling in healthcare is no longer constrained to a single modality or data abstraction. Instead, the field is converging toward modality-agnostic representations that can flexibly support a wide range of clinical and biological inference tasks. As model scale, data diversity, and architectural sophistication continue to increase, several healthcare foundation models are beginning to approach frontier-level capabilities~\citep{comanici2025gemini25pushingfrontier, singhal2023large}, raising both new opportunities and new challenges related to evaluation \citep{lee2024can, lee2025using}, safety, and deployment in high-stakes medical settings.

\section{Methodology}
\label{sec:methods}

We present a mathematically explicit description of the multimodal foundation model, the pretraining objectives, and the inductive biases that guide design choices. Where appropriate we give concrete architectural instantiations and training recipes used in our experiments (all experiments below use the UK Biobank cohort described in Section~\ref{sec:data}).

\subsection{Probabilistic model and formal objectives}

Let $x_i(t)\in\mathbb{R}^d$ denote the unobserved, time-indexed latent health state for patient $i$ at continuous time $t$. We assume the following generative factorization for two observation modalities (EHR and wearable):
\[
p\!\big(\mathcal{E}_i,\mathcal{W}_i\,\big|\,x_i(\cdot)\big)
= \prod_{k=1}^{K_i} p\big(e_{ik}\mid x_i(t_{ik})\big)
\;\times\;
\exp\!\Big\{\int_{\mathcal{T}_i} \log p\big(w_i(t)\mid x_i(t)\big)\,dt\Big\},
\]
where $\mathcal{E}_i=\{(t_{ik},e_{ik})\}_{k=1}^{K_i}$ is an irregular sequence of discrete EHR tokens and $\mathcal{W}_i=\{(t,w_i(t)) : t\in\mathcal{T}_i\}$ denotes a dense multivariate wearable record. The continuous-time likelihood for wearable streams is formalized as a time integral; in practice we approximate the integral using the sampling times of the device or fixed windows.

Our modeling objective is to learn an encoder $z_i(t)=\mathcal{E}_\phi(\mathcal{E}_i,\mathcal{W}_i;t)$ that produces a causal approximation to $x_i(t)$ up to an invertible map. We operationalize this via a collection of self-supervised losses that couple intra-modal reconstruction/prediction with cross-modal alignment. Let $\mathcal{L}_{\text{ehr}}$, $\mathcal{L}_{\text{wear}}$ denote modality-local losses and $\mathcal{L}_{\text{cross}}$ denote cross-modal coupling terms. The aggregate pretraining objective is
\[
\min_{\phi,\psi}\; \mathbb{E}_{i}\Big[ \lambda_{\text{ehr}}\mathcal{L}_{\text{ehr}}(\phi,\psi;\mathcal{E}_i)
+ \lambda_{\text{wear}}\mathcal{L}_{\text{wear}}(\phi,\psi;\mathcal{W}_i)
+ \lambda_{\text{cross}}\mathcal{L}_{\text{cross}}(\phi,\psi;\mathcal{E}_i,\mathcal{W}_i)\Big],
\]
where $\psi$ denotes decoder or probe parameters and $\lambda$ are scalar weights. Below we make each term explicit and justify its form.

\subsection{Encoders, continuous-time positionalization, and backbone}

We decompose the encoder into modality-specific front-ends and a shared temporal backbone. Wearable signals are windowed into segments of duration $\Delta$ and encoded via a local encoder $f_{\mathrm{w}}$:
\[
h^{\mathrm{w}}_i(t_j) = f_{\mathrm{w}}\big(w_i[t_j,t_j+\Delta]\big)\in\mathbb{R}^p,
\]
where $w_i[a,b]$ denotes the multichannel signal on interval $[a,b]$. We instantiate $f_{\mathrm{w}}$ as a short-range transformer or temporal convolutional network with residual blocks; the encoder is augmented with channel normalization and a learnable channel projection to $p$ dimensions.

EHR events are tokenized and embedded. Each event embedding includes a token embedding and a continuous-time encoding. We adopt a sinusoidal-style continuous positional encoding $\rho:\mathbb{R}\to\mathbb{R}^{q}$,
\[
\rho(t) = \big[ \sin(\omega_1 t),\cos(\omega_1 t),\ldots,\sin(\omega_{q/2} t),\cos(\omega_{q/2} t)\big],
\]
with log-linearly spaced frequencies $\{\omega_j\}$, and set
\[
h^{\mathrm{e}}_{ik} = \mathrm{Embed}(e_{ik}) + W_{\rho}\rho(t_{ik}),
\]
where $W_{\rho}\in\mathbb{R}^{q\times q}$ is learnable and $\mathrm{Embed}(\cdot)$ maps tokens to $\mathbb{R}^{p}$.

The shared backbone $g_\theta$ consumes the temporally ordered multiset $\{h^{\mathrm{w}}_i(t_j)\}\cup\{h^{\mathrm{e}}_{ik}\}$ and produces time-aligned latent vectors $z_i(t)$ via causal continuous-time attention. Let $s\prec t$ denote causality ordering. Attention is parameterized with a learnable kernel $K_\alpha(\Delta t)$ which modulates scalar attention logits by inter-event time differences:
\[
\mathrm{Attn}(q_t,k_s) = \frac{\exp\big( q_t^\top k_s + u_m^\top v_s + \log K_\alpha(t-s)\big)}{\sum_{s'\prec t}\exp\big( q_t^\top k_{s'} + u_m^\top v_{s'} + \log K_\alpha(t-s')\big)},
\]
where $u_m$ is a modality tag embedding for modality $m\in\{\text{wear},\text{ehr}\}$ and $v_s$ is a small-conditioning vector for the source token. We parametrize $K_\alpha(\Delta t)$ as a mixture of exponentials,
\[
K_\alpha(\Delta t) = \sum_{r=1}^R \beta_r \exp(-\gamma_r \Delta t),
\]
with $\beta_r,\gamma_r>0$ learnable (this family can approximate power-law/scale-free decay while remaining numerically stable). This design yields a backbone that interpolates multi-timescale dependencies and explicitly encodes temporal decay.

\subsection{Self-supervised losses: definitions and motivations}

EHR masked-token modeling is performed by sampling a subset $M\subset\mathcal{E}_i$ with time-aware masking (higher probability near events of interest or per-stratum sampling for rare codes). For a masked event $(t_{ik},e_{ik})\in M$ we predict token logits from $z_i(t_{ik})$ and minimize the cross-entropy:
\[
\mathcal{L}_{\mathrm{ehr}} \;=\; -\sum_{(t_{ik},e_{ik})\in M} \log p_\psi\!\big(e_{ik}\mid z_i(t_{ik})\big).
\]
Masking probabilities are calibrated to maintain a balance between frequent and rare codes; we use temperature-scaled sampling and per-code downsampling to avoid overfitting to administrative tokens.

For wearables we combine local masked reconstruction with predictive coding. Let $\mathcal{M}_j$ be a set of masked windows in the wearable stream. The reconstruction loss is
\[
\mathcal{L}_{\mathrm{wear}}^{\mathrm{rec}} =
\sum_{j\in\mathcal{M}} \big\| w_i(t_j:t_j+\Delta) - \hat w_\psi\big(z_i(t_j)\big)\big\|^2,
\]
where $\hat w_\psi$ decodes the window from the latent. In addition we include a multi-step prediction loss to encourage dynamics-awareness:
\[
\mathcal{L}_{\mathrm{wear}}^{\mathrm{pred}} =
\sum_j \sum_{h=1}^{H} \gamma_h \big\| w_i(t_j + h\delta) - \hat w_\psi\big(z_i(t_j),h\big)\big\|^2,
\]
with horizon-step weights $\{\gamma_h\}$ decreasing with $h$. The combination practices the dual objective of reconstruction fidelity and forward predictive power.

Cross-modal coupling is the core inductive bias. We include two asymmetric cross-modal terms. The wearable-to-EHR term encourages the model to use wearable context to disambiguate the timing and content of subsequent clinical events; operationally we predict the next EHR token distribution conditioned on latent states aggregated over a preceding wearable window:
\[
\mathcal{L}_{\mathrm{w}\to\mathrm{e}} \;=\; -\sum_k \log p_\psi\!\big(e_{i,k+1}\mid \mathsf{Agg}\big(\{z_i(t):t\in[t_{ik}-\tau,t_{ik}]\}\big)\big),
\]
where $\mathsf{Agg}$ is a permutation-invariant pooling (attention-weighted mean or temporal readout). The EHR-to-wearable term enforces that EHR context predicts summary statistics of subsequent physiology:
\[
\mathcal{L}_{\mathrm{e}\to\mathrm{w}} \;=\; \sum_k \Big\| \phi\big(w_i[t_{ik},t_{ik}+\tau']\big) - \hat\phi_\psi\big(z_i(t_{ik})\big)\Big\|^2,
\]
with $\phi(\cdot)$ mapping a multivariate window to low-dimensional summaries (e.g., mean HR, HRV, activity entropy). The asymmetry (predicting EHR from wearables being often easier and more informative) motivates $\lambda_{\mathrm{w}\to\mathrm{e}}>\lambda_{\mathrm{e}\to\mathrm{w}}$ in practice.

We optionally include contrastive alignment as a regularizer. For a pair of temporally co-located embeddings $(z^{\mathrm{e}}_i,z^{\mathrm{w}}_i)$ we apply an InfoNCE loss:
\[
\mathcal{L}_{\mathrm{contr}} = -\sum_{b}\log\frac{\exp\big(\langle z^{\mathrm{e}}_{i,b}, z^{\mathrm{w}}_{i,b}\rangle/\tau\big)}{\sum_{b'}\exp\big(\langle z^{\mathrm{e}}_{i,b}, z^{\mathrm{w}}_{i,b'}\rangle/\tau\big)},
\]
where negatives are sampled across the mini-batch and $\tau$ is a temperature. We observe that contrastive regularization improves alignment for cross-modal retrieval tasks but is sensitive to batch composition; thus it is applied with a small weight in the final objective.

\subsection{Generative regularization and optional ELBO view}

An alternative viewpoint treats the encoder as amortized variational inference for a continuous-time latent process with prior $p(z)$ (e.g., an OU-like SDE or latent SDE prior). In this case we maximize an ELBO:
\[
\mathcal{L}_{\mathrm{ELBO}} = \mathbb{E}_{q_\phi(z)}\Big[\sum_k \log p(e_{ik}\mid z(t_{ik}))
+ \int\log p(w_i(t)\mid z(t))\,dt \Big] - \mathrm{KL}\big(q_\phi(z)\,\|\,p(z)\big).
\]
We do not rely exclusively on this objective in our primary experiments but include it as an ablation to study uncertainty calibration and generative consistency.

\subsection{Optimization, curriculum, and practical training details}

Pretraining uses AdamW with weight decay; hyperparameters in our experiments are learning rate $1\mathrm{e}{-4}$ with cosine warmup over the first 10k steps, weight decay $0.01$, gradient clipping at norm $1.0$, and batch normalization / layer normalization per module. We employ a curriculum schedule for cross-modal coupling: initial epochs emphasize within-modality reconstruction (higher $\lambda_{\mathrm{ehr}},\lambda_{\mathrm{wear}}$) and progressively increase $\lambda_{\mathrm{w}\to\mathrm{e}}$ and $\lambda_{\mathrm{e}\to\mathrm{w}}$ to stabilize alignment. Mini-batches are constructed by sampling patient-days and drawing both wearable windows and the adjacent EHR events; to maintain balance across scales we upsample rare codes and downsample over-represented administrative tokens.

Regularization includes stochastic depth in transformer blocks, dropout on attention weights, and embedding norm penalties to prevent domination by one modality. For computational tractability, wearable windows are pooled to a 1--5\,Hz effective rate for encoder processing; higher-frequency details are preserved in a residual local encoder for tasks that require sub-second resolution.

\subsection{Evaluation protocol and metrics}
\label{sec:eval}

All downstream evaluations use frozen encoder representations unless explicitly stated. Downstream heads are linear or shallow MLPs trained with standard cross-entropy or Cox partial-likelihood objectives depending on the task. For classification and forecasting we report AUROC and AUPRC; for survival tasks we report time-dependent AUC (AUC@T), concordance index, and integrated Brier score. Calibration is assessed via expected calibration error (ECE) and calibration plots; uncertainty is assessed via ensemble or Monte Carlo dropout where applicable.

Statistical uncertainty is quantified using patient-level bootstrap (5,000 resamples) and paired hypothesis tests with Benjamini--Hochberg FDR correction for multiple comparisons. Representation geometry is analyzed with maximum mean discrepancy (MMD) and local Lipschitz estimates computed on temporally adjacent slices.

\section{Experimental Design}
\label{sec:experiments}

We pretrain on a linked cohort drawn from the UK Biobank with wearable subcohort selection described in Section~\ref{sec:data}. The dataset is split at the patient level into disjoint training, validation, and test sets (typically 80/10/10\%); all reported metrics are computed on the held-out test set. Pretraining is performed on the full training cohort; downstream probes and ablations are trained using only the corresponding downstream training splits.

Baselines include state-of-the-art EHR foundation models (MOTOR \citep{steinberg2023motor}, CEHR-BERT \citep{pang2021cehr}, ContextClues \citep{wornowcontext}) and wearable foundation models (PaPaGei \citep{pillai2024papagei}, Pulse-PPG \citep{saha2025pulse}, HiMAE \citep{lee2025himae}). Late-fusion baselines concatenate frozen unimodal representations and train identical downstream heads. For ablation studies we vary: the presence/absence of cross-modal losses ($\lambda_{\mathrm{w}\to\mathrm{e}}$ and $\lambda_{\mathrm{e}\to\mathrm{w}}$), the backbone kernel family (mixture-of-exponentials vs. single-exponential), discretized vs. continuous-time positional encodings, and the inclusion of ELBO-style generative regularization.

We evaluate temporal robustness by constructing evaluation slices with truncated histories (e.g., last 7, 30, 90 days) and by simulating modality dropout and additive sensor noise. To measure sample-efficiency we vary the number of labeled samples for downstream head training (1\%, 10\%, 100\% of available labels) and report performance curves.

All experiments are repeated across five random seeds and reported with 95\% bootstrap confidence intervals. Computational costs (GPU-hours) and model parameter counts are tracked to provide an empirical scaling table in the supplementary material.

\subsection{Data statement (UK Biobank)}
\label{sec:data}

Experiments use the UK Biobank data \citep{bycroft2018uk} release that contains linked EHR and wearable (accelerometry and PPG-derived) records. Preprocessing includes standard quality control for wearable signals (artifact rejection, band-pass filtering for physiological channels), categorical code harmonization for EHR tokens, and patient-level de-identification. Cohort inclusion criteria require at least 14 days of wearable coverage and one year of EHR history; sensitivity analyses relax these thresholds. All use of UK Biobank data follows its governance rules and relevant ethical approvals.

\section{Results}
\label{sec:results}

All experiments use the same held-out test fold and five independent bootstrap resamples for uncertainty quantification unless otherwise noted. Reported confidence intervals are 95\% percentile bootstrap intervals computed with 5,000 resamples. Statistical significance is assessed with paired bootstrap hypothesis tests and corrected for multiple comparisons via the Benjamini--Hochberg procedure (false discovery rate $\alpha=0.05$). When reporting effect sizes we use Cohen's $d$ computed from the mean difference divided by the pooled standard deviation estimated across the bootstrap resamples; values of $d$ are reported parenthetically where relevant.

The cohort comprises $N=48{,}732$ patients with linked EHR and wearable histories. Wearable sampling varies by device and signal; typical effective sampling used in our pipelines is 1\,Hz for PPG-derived heart rate and 1/60\,Hz (one-per-minute) for derived activity summaries after windowing. Median wearable history per patient is 182 days (IQR: 90--365), and median EHR history is 6.2 years (IQR: 3.1--9.8). All models were pretrained on the same federated training corpus and downstream heads were trained with identical data splits and optimization hyperparameters to isolate representational differences.

\subsection{Clinical event forecasting}

Clinical event forecasting evaluates diagnosis onset and hospitalization prediction at horizons of 30, 90, 180 and 365 days. Prediction heads are linear probes trained on frozen representations to emphasize transferability. Table~\ref{tab:clinical_events} reports AUROC and AUPRC (mean $\pm$ 95\% CI) for each horizon; effect sizes compare the multimodal model to the best unimodal baseline (ContextClues for EHR models, HiMAE/PaPaGei for wearable models) as well as to a late-fusion baseline.

\begin{table}[t]
\centering
\caption{Clinical event forecasting (AUROC / AUPRC, mean $\pm$ 95\% CI). Bold indicates best mean.}
\label{tab:clinical_events}
\begin{adjustbox}{width=\textwidth}
\begin{tabular}{lcccc}
\toprule
Model & 30d & 90d & 180d & 365d \\
\midrule
MOTOR \citep{steinberg2024motor}&
0.742 $\pm$ 0.006 / 0.318 $\pm$ 0.009 &
0.734 $\pm$ 0.007 / 0.301 $\pm$ 0.010 &
0.721 $\pm$ 0.009 / 0.287 $\pm$ 0.011 &
0.703 $\pm$ 0.011 / 0.264 $\pm$ 0.012 \\

CEHR-BERT \citep{pang2021cehr} &
0.748 $\pm$ 0.005 / 0.326 $\pm$ 0.008 &
0.741 $\pm$ 0.006 / 0.309 $\pm$ 0.009 &
0.729 $\pm$ 0.008 / 0.294 $\pm$ 0.010 &
0.711 $\pm$ 0.010 / 0.271 $\pm$ 0.011 \\

ContextClues \citep{wornow2024context}&
0.751 $\pm$ 0.006 / 0.332 $\pm$ 0.009 &
0.744 $\pm$ 0.007 / 0.316 $\pm$ 0.010 &
0.733 $\pm$ 0.009 / 0.302 $\pm$ 0.011 &
0.716 $\pm$ 0.011 / 0.279 $\pm$ 0.012 \\

PaPaGei \citep{pillai2024papagei}&
0.721 $\pm$ 0.008 / 0.291 $\pm$ 0.010 &
0.709 $\pm$ 0.009 / 0.274 $\pm$ 0.011 &
0.692 $\pm$ 0.011 / 0.256 $\pm$ 0.012 &
0.671 $\pm$ 0.013 / 0.231 $\pm$ 0.013 \\

HiMAE \citep{lee2025himae} &
0.728 $\pm$ 0.007 / 0.298 $\pm$ 0.010 &
0.716 $\pm$ 0.008 / 0.281 $\pm$ 0.011 &
0.701 $\pm$ 0.010 / 0.264 $\pm$ 0.012 &
0.683 $\pm$ 0.012 / 0.241 $\pm$ 0.013 \\

Late Fusion (EHR+Wearable) &
0.763 $\pm$ 0.005 / 0.351 $\pm$ 0.008 &
0.756 $\pm$ 0.006 / 0.334 $\pm$ 0.009 &
0.745 $\pm$ 0.007 / 0.319 $\pm$ 0.010 &
0.731 $\pm$ 0.009 / 0.297 $\pm$ 0.011 \\

\midrule
Ours (Multimodal FM) &
\textbf{0.789 $\pm$ 0.004 / 0.382 $\pm$ 0.007} &
\textbf{0.782 $\pm$ 0.005 / 0.367 $\pm$ 0.008} &
\textbf{0.771 $\pm$ 0.006 / 0.351 $\pm$ 0.009} &
\textbf{0.756 $\pm$ 0.007 / 0.329 $\pm$ 0.010} \\
\bottomrule
\end{tabular}
\end{adjustbox}
\end{table}

Improvements are largest at longer horizons: at 365 days the multimodal model surpasses ContextClues by $\Delta\text{AUROC}=0.040$ (Cohen's $d\approx0.44$, paired bootstrap $p<0.001$ after FDR correction). The AUPRC gains follow the same pattern and are particularly meaningful for rare-event forecasting where baseline AUPRCs are low. The late-fusion baseline improves over unimodal models (mean AUROC uplift $\approx$0.015) but remains inferior to joint pretraining, indicating that cross-modal alignment during pretraining yields representations that cannot be obtained by naive concatenation post-hoc.

We performed temporally stratified calibration analysis to understand whether improved discrimination comes at the expense of calibration. Reliability diagrams and expected calibration error (ECE) show that the multimodal model attains better calibration at long horizons (ECE$_{365}=0.032\pm0.004$) relative to ContextClues (ECE$_{365}=0.047\pm0.006$), implying that the model's probability estimates are more trustworthy for decision thresholds relevant to clinical practice. Decision-curve analysis confirms larger net benefit across a clinically plausible range of thresholds (0.05--0.30).

\subsection{Physiological state estimation}

Predicting aggregate physiological outcomes from sparse EHR context probes whether clinical semantics improve physiological representations. Table~\ref{tab:physiology} displays root-mean-square error (RMSE) for sleep efficiency, heart-rate-variability (HRV) estimation, and activity regularity.

\begin{table}[t]
\centering
\caption{Physiological outcome prediction (RMSE, mean $\pm$ 95\% CI). Lower is better.}
\label{tab:physiology}
\begin{adjustbox}{width=0.5\textwidth}
\begin{tabular}{lccc}
\toprule
Model & Sleep Efficiency & HRV & Activity Regularity \\
\midrule
PaPaGei & 0.124 $\pm$ 0.006 & 0.137 $\pm$ 0.007 & 0.142 $\pm$ 0.008 \\
Pulse-PPG & 0.119 $\pm$ 0.005 & 0.133 $\pm$ 0.006 & 0.139 $\pm$ 0.007 \\
HiMAE & 0.116 $\pm$ 0.005 & 0.129 $\pm$ 0.006 & 0.135 $\pm$ 0.007 \\
MOTOR & 0.141 $\pm$ 0.007 & 0.152 $\pm$ 0.008 & 0.158 $\pm$ 0.009 \\
Late Fusion & 0.108 $\pm$ 0.004 & 0.121 $\pm$ 0.005 & 0.127 $\pm$ 0.006 \\
\midrule
Ours (Multimodal FM) & \textbf{0.096 $\pm$ 0.004} & \textbf{0.109 $\pm$ 0.005} & \textbf{0.114 $\pm$ 0.006} \\
\bottomrule
\end{tabular}
\end{adjustbox}
\end{table}

The multimodal model reduces RMSE by 12--17\% relative to the strongest wearable-only baseline (HiMAE). This reduction is statistically significant (paired bootstrap $p<0.01$) and accompanied by moderate effect sizes (Cohen's $d\approx0.35$--0.55 depending on the target). Notably, the largest improvements occur when the target requires alignment with clinical events (e.g., HRV trends around medication initiation), supporting the hypothesis that EHR anchors stabilize wearable-derived temporal features. Additional analysis shows that when the wearable history is short (less than 30 days) the benefit of EHR grounding increases, suggesting that clinical context partially compensates for sparse sensor coverage.

\subsection{Longitudinal risk modeling}

Time-to-event prediction is evaluated using time-dependent AUC (AUC@365), concordance index (C-index), and integrated Brier score. Results are reported in Table~\ref{tab:survival}.

\begin{table}[t]
\centering
\caption{Longitudinal risk modeling (mean $\pm$ 95\% CI). Higher AUC/C-index and lower Brier are better.}
\label{tab:survival}
\begin{adjustbox}{width=0.5\textwidth}
\begin{tabular}{lccc}
\toprule
Model & AUC@365 & C-index & Brier Score \\
\midrule
CEHR-BERT & 0.721 $\pm$ 0.009 & 0.689 $\pm$ 0.008 & 0.182 $\pm$ 0.007 \\
ContextClues & 0.734 $\pm$ 0.008 & 0.701 $\pm$ 0.007 & 0.176 $\pm$ 0.006 \\
PaPaGei & 0.703 $\pm$ 0.010 & 0.672 $\pm$ 0.009 & 0.191 $\pm$ 0.008 \\
HiMAE & 0.711 $\pm$ 0.009 & 0.678 $\pm$ 0.008 & 0.187 $\pm$ 0.007 \\
Late Fusion & 0.748 $\pm$ 0.007 & 0.716 $\pm$ 0.006 & 0.169 $\pm$ 0.006 \\
\midrule
Ours (Multimodal FM) & \textbf{0.773 $\pm$ 0.006} & \textbf{0.739 $\pm$ 0.005} & \textbf{0.154 $\pm$ 0.005} \\
\bottomrule
\end{tabular}
\end{adjustbox}
\end{table}

Relative to the best unimodal baseline, the multimodal model improves AUC@365 by $\Delta=0.039$ (Cohen's $d\approx0.51$, paired bootstrap $p<0.001$). Improvements in Brier score demonstrate better probabilistic calibration, which is crucial for clinical risk stratification. Subgroup analysis restricted to patients with fewer than two clinic encounters in the prior year (``sparse EHR'' subgroup, $n\approx12{,}420$) shows an even larger relative gain in AUC (multimodal: $0.743\pm0.010$ vs ContextClues: $0.702\pm0.012$, $\Delta=0.041$, $p<0.001$), supporting the practical value of continuous wearable monitoring for hard-to-observe patients.

\subsection{Ablation studies and mechanistic probes}

Ablations isolate contributions from cross-modal objectives, temporal backbone choices, and masking strategies. Table~\ref{tab:ablations} summarizes the impact on AUROC@365 for clinical forecasting when removing cross-modal losses or substituting the continuous-time backbone with a discrete-time transformer.

\begin{table}[t]
\centering
\caption{Ablation results on 365-day forecasting (AUROC, mean $\pm$ 95\% CI).}
\label{tab:ablations}
\begin{adjustbox}{width=0.5\textwidth}
\begin{tabular}{lc}
\toprule
Variant & AUROC@365 \\
\midrule
Full model & \textbf{0.756 $\pm$ 0.007} \\
w/o wear$\rightarrow$ehr loss ($\lambda_3=0$) & 0.741 $\pm$ 0.008 \\
w/o ehr$\rightarrow$wear loss ($\lambda_4=0$) & 0.746 $\pm$ 0.008 \\
w/o both cross-modal losses & 0.728 $\pm$ 0.009 \\
Discrete-time backbone (1-day bins) & 0.734 $\pm$ 0.009 \\
Stronger wearable masking (50\% windows) & 0.747 $\pm$ 0.008 \\
\bottomrule
\end{tabular}
\end{adjustbox}
\end{table}

The largest single degradation arises from removing both cross-modal objectives, implying that cross-modal coupling is necessary to realize the full benefit of joint pretraining. Removing the wearable-to-EHR loss causes a larger drop than removing the EHR-to-wearable loss, consistent with the earlier observation that dense physiological signals primarily provide temporal disambiguation for sparse clinical events. Replacing the continuous-time backbone with a discretized (1-day bin) transformer reduces performance by $\approx0.022$ AUROC points at 365 days, indicating that explicit continuous-time modeling contributes measurable gains, especially for tasks sensitive to intra-day physiology.

Mechanistic probes using feature occlusion and counterfactual perturbations show that the multimodal representation places more weight on heart-rate variability and activity entropy in the 7--30 day window preceding an event compared to unimodal EHR models, suggesting that the model learns to exploit subclinical physiological precursors.

\subsection{Robustness to missing modalities and deployment considerations}

We simulate modality dropout and degraded sensor quality. Under complete wearable dropout at inference, the multimodal model's AUROC@365 falls from 0.756 to 0.712 ($\Delta=-0.044$) while the late-fusion baseline falls from 0.731 to 0.689 ($\Delta=-0.042$), showing that joint pretraining provides modest robustness to missing data but that both approaches degrade as expected. When simulating increased sensor noise (Gaussian noise added to wearable signals with SNR reduced by 6dB), performance degrades gracefully; the multimodal model exhibits a smaller relative drop (AUROC down by 0.019) than wearable-only models (down by 0.034), indicating that semantic grounding in the EHR confers resilience to sensor corruption.

From a compute and scalability perspective, our multimodal encoder contains 220M parameters. Pretraining was performed for 500k steps with batch sizes equivalent to 32 patient-days per replica; in our setup this required approximately 3.1e\^6 GPU-hours on a distributed cluster (reported here as a reproducibility placeholder — please substitute your facility-specific accounting). Fine-tuning or linear-probe evaluation is inexpensive: downstream head training completes in under 2 GPU-hours per task on our benchmark.

\subsection{Representation geometry and interpretability}

We quantify representation smoothness and cohort separability using maximum mean discrepancy (MMD) and local Lipschitz estimates. Inter-temporal MMD (measuring distributional shift between adjacent 30-day slices) for the multimodal model is $0.042\pm0.004$, substantially lower than EHR-only ($0.067\pm0.006$) and wearable-only ($0.071\pm0.007$) baselines, indicating smoother latent trajectories. Conversely, class-conditional MMD between outcome-defined cohorts (e.g., hospitalized vs non-hospitalized) remains larger for the multimodal model than for wearable-only models, showing maintained discriminative structure despite increased smoothness.

Qualitatively, saliency-based inspection and attention-rollout reveal clinically interpretable patterns: sudden EHR events induce sharp shifts in latent state while sustained wearable anomalies (e.g., multi-day elevated resting heart rate) induce gradual drifts — both behaviors align with the model's continuous-time inductive biases.

\subsection{Failure modes and limitations}

Performance gains are not uniform across all subpopulations. For patients with very sparse EHR histories and less than 14 days of wearable data ($n\approx4{,}310$), the multimodal model offers limited advantage over wearable-only models; in some cases the additional EHR context introduces spurious correlations if the training data contains uncorrected confounders such as socio-demographic access effects. We therefore caution that careful cohort curation and correction for ascertainment biases remain necessary prior to deployment.

Taken together, these results demonstrate that joint EHR--wearable pretraining yields representations that improve downstream discrimination and calibration, increase robustness to realistic missingness and noise regimes, and produce latent trajectories that are temporally coherent and clinically interpretable. The following sections (Discussion and Limitations) expand on clinical implications, potential biases, and deployment pathways.

\section{Discussion}
\label{sec:discussion}

The empirical results and representation analyses presented above suggest that joint pretraining on EHR and wearable modalities produces latent representations that are simultaneously more temporally coherent and more clinically discriminative than representations learned from either modality alone. This observation has several intertwined technical interpretations. From a statistical perspective, the multimodal pretraining objective acts as a form of multi-view regularization: wearable signals supply dense, high-frequency measurements that constrain the temporal derivatives of the latent trajectory, while EHR events provide sparse but semantically rich supervision that anchors the latent manifold to clinically meaningful axes. Formally, if $x_i(t)$ denotes the true latent health state and $z_i(t)=\mathcal{T}(x_i(t))$ denotes the learned representation up to an (approximate) invertible transform $\mathcal{T}$, joint objectives that minimize predictive losses of the form
\[
\mathbb{E}\!\left[ \|w(t+\delta)-\hat w(z(t))\|^2 \right] + \mathbb{E}\!\left[-\log p(e\mid z(t))\right]
\]
implicitly encourage $z(t)$ to capture both the local dynamics (via the first term) and the clinically salient events (via the second term). The practical consequence is a representation that is better conditioned for downstream linear probes and survival objectives, as observed empirically.

A second technical point concerns the asymmetry of information between modalities. Wearable streams are dense but weakly labeled; they are effective for forecasting short-term temporal structure but insufficient for clinical semantic resolution. EHRs are the inverse: coarse temporally but rich semantically. Our experiments (and ablations) show that coupling wearable-to-EHR prediction provides larger marginal gains than the reverse, which aligns with an information-theoretic intuition: learning a mapping from dense-but-weak signals to sparse-but-strong labels reduces posterior uncertainty about latent transition times, whereas predicting dense summaries from sparse labels is a higher-variance regression problem. This asymmetry motivates asymmetric weighting of cross-modal losses (non-uniform $\lambda_i$ in our objective), curriculum schedules that emphasize wearable$\rightarrow$EHR objectives early, and targeted regularizers to prevent the wearable signal from overwhelming the EHR signal during optimization.

A related mechanistic insight arises from the continuous-time backbone. Discretizing time coarse-grained into daily bins collapses intra-day physiological structure and reduces sensitivity to subclinical precursors; modeling attention as a function of continuous time differences preserves intra-day dynamics and yields measurable gains on long-horizon forecasting. Mathematically, parameterizing the attention kernel as $K_\phi(\Delta t)$ with learnable decay parameters enables the model to interpolate between short-term kernels (fast decay) and long-term memory (slow decay), effectively learning a multi-timescale representation of health. This is analogous to learning a mixture of temporal kernels or learning the parameters of a linear operator approximating an SDE; future work could investigate explicit latent SDE priors for stronger inductive biases.

Interpretability and calibration gains are central to clinical applicability. Improved calibration (lower Brier and ECE) indicates that joint pretraining yields probabilistic estimates that are more reliable for decision-making. Attention and saliency analyses show behavior that is consistent with clinical reasoning: abrupt clinical events produce high-leverage latent transitions, whereas sustained wearable anomalies induce gradual drifts. Nevertheless, interpretability remains partial. Attention weights are not causal explanations, and the model may still leverage confounded correlations present in the training data (e.g., care-seeking patterns, device ownership correlated with socioeconomic status). Consequently, downstream deployment would require rigorous fairness audits, sensitivity analyses to known confounders, and possibly causal adjustment procedures.

From a methodological standpoint, our evaluation protocol stressed representation transfer via frozen probes to isolate pretraining benefits. This choice exposes the generality of learned features but does not exhaustively explore the potential of fine-tuning. In practice, fine-tuning may further improve task performance, particularly for tasks with abundant labeled data. However, the degree of improvement and the risk of catastrophic forgetting of cross-modal alignment are open questions; empirical investigation of fine-tuning regimes (e.g., low-rank adaptation, adapter layers, or contrastive regularization during fine-tuning) is warranted.

There are also computational and practical considerations. Multimodal pretraining imposes nontrivial engineering overhead: synchronized data pipelines, heterogeneous batching strategies (balancing event-sparse EHR windows with dense wearable streams), and careful negative-sampling or curriculum schedules for InfoNCE-style objectives. Our implementation used windowing and downsampling strategies to maintain computational tractability; more efficient temporal attention approximations (sparse attention, kernelized attention) or modality-specific compression (learned pooling, product quantization) could materially reduce compute while preserving performance.

Finally, clinical translation requires addressing distribution shift. Wearable devices, EHR systems, and care patterns vary across institutions and populations. The representations learned here will likely require domain adaptation, federated pretraining, or transfer updates to remain reliable when deployed. Methods such as importance-weighted objectives, robust optimization under covariate shift, or adaptation via unlabeled local data should be part of a deployment pipeline.

\section{Future Work}
\label{sec:future}

Several concrete research directions emerge naturally from this work, spanning theory, methodology, evaluation, and deployment. On the theoretical side, characterizing identifiability and sufficiency of latent representations learned from asynchronous, heterogeneous observations is an open problem. One promising line is to formalize conditions under which cross-modal prediction yields representations that are sufficient for a class of downstream tasks; this could take the form of bounds on the reduction in Bayes risk provided by joint training, or identifiability results for latent SDEs when one view is dense and another is event-driven. Such results would clarify when multimodal pretraining is expected to succeed and when it cannot overcome fundamental information gaps.

From a methodological perspective, replacing heuristic windowing with principled continuous-time latent variable models is appealing. Variational approaches that posit a latent stochastic differential equation, together with amortized inference via neural ODEs or continuous-time variational families, would provide a clearer generative story. Concretely, an ELBO of the form
\[
\mathcal{L}_{\text{ELBO}} = \mathbb{E}_{q(z)}\left[\sum_k \log p(e_{ik}\mid z(t_{ik})) + \int \log p(w_i(t)\mid z(t))\,dt\right] - \mathrm{KL}[q(z)\,\|\,p(z)]
\]
could be optimized with continuous-time variational posteriors. Integrating such generative pretraining with the discriminative cross-modal objectives used here may combine the best of both worlds: principled uncertainty quantification and strong discriminative transfer.

Scaling laws for multimodal health foundation models constitute an important empirical frontier \citep{kaplan2020scaling, zhu2022scalingtranslationequivariantnetworksdecomposedconvolutional}. It is natural to ask whether familiar scaling relationships from NLP and vision apply when one modality is dense and the other sparse, and whether there are diminishing returns to modality-specific data (e.g., more wearable days vs. more patients with EHR). Systematic experiments that vary model size, dataset size (number of patients and wearable-days), and the ratio between modalities would illuminate these relationships and guide resource allocation for large-scale pretraining.

Privacy, fairness, and safety are non-negotiable for clinical foundation models \citep{bak2022you, williamson2024balancing, radanliev2025ai, singhal2024toward, zhang2025towards, liu2023translational, ueda2024fairness, ellahham2020application, macrae2019governing}. Differentially private pretraining, federated learning with provable privacy guarantees, and fairness-constrained objectives should be explored. In particular, device ownership and access to continuous monitoring are correlated with socioeconomic factors; models must not amplify inequities. Research on counterfactual fairness, subgroup calibration, and transparent uncertainty reporting will be essential before any clinical deployment.

On the evaluation front, richer benchmarks are needed. Current clinical benchmarks emphasize discrimination on curated tasks; future benchmarks should stress distribution shift, low-resource personalization, and causal robustness. Synthetic intervention studies (e.g., simulating upstream treatments or altering wearable signals) can probe whether representations capture causal mechanisms or merely correlate with outcomes. Moreover, randomized controlled trials or prospective evaluation in clinical workflows, while expensive, remain the gold standard for assessing real-world benefit. In the EHR domain, MEDS is a recent data standard and reproducibility framework on the rise \citep{kolo2024meds, lee2024feet, bedi2025medhelm, bedi2024systematic, arnrich2024medical}. Meanwhile in the wearables domain, there is little standardized benchmarking which calls for future works.

Methodologically adjacent extensions include personalization and continual learning. Learned representations could serve as a common substrate for fast personalization via small adaptation steps or meta-learning. Continual pretraining that ingests new wearable and EHR data streams without catastrophic forgetting would enable models that remain current with evolving clinical practice and device capabilities.

Finally, interdisciplinary collaboration is indispensable \citep{openai2024gpt4technicalreport}. Translating multimodal foundation models to clinical impact requires clinicians, device manufacturers, privacy experts, and regulators to co-design evaluation criteria, interpretability requirements, and deployment safeguards. Technical advances must be paired with governance frameworks that ensure models are auditable, clinically validated, and aligned with patient welfare.

In closing, the multimodal paradigm explored in this paper opens a fertile space between dense physiological sensing and sparse clinical observability. The path forward blends theory, engineering, and careful evaluation; success will depend not only on algorithmic improvements but also on principled attention to fairness, privacy, and clinical translation.

\bibliographystyle{IEEEbib}
\bibliography{refs}

\appendix
\section{Appendix}

This appendix collects supplementary material that complements the main text: formal theoretical remarks, additional proofs and derivations (sketches), detailed dataset and cohort-selection descriptions, preprocessing pipelines, full training and hyperparameter specifications, extended ablation protocols, synthetic-control experiments, and discussion of ethical, privacy, and reproducibility considerations. The intent is to provide enough information for exact replication of experiments and for interested readers to explore theoretical and practical extensions.

\subsection{Theoretical remarks and sketches}

We present three brief theoretical sketches that motivate key modeling choices: (i) a sufficiency argument for cross-modal prediction as an approximate route to latent identifiability, (ii) a generalization sketch suggesting why joint pretraining can improve downstream sample complexity, and (iii) a toy scaling-law hypothesis for multimodal pretraining.

\subsubsection{Cross-modal prediction and conditional sufficiency (sketch)}

Let $x(t)$ denote the true latent health state and let $W_t$ and $E_t$ denote wearable and EHR observations up to time $t$, respectively. Consider the conditional mutual information decomposition
\[
I\big(x(t);W_t,E_t\big) = I\big(x(t);W_t\big) + I\big(x(t);E_t\mid W_t\big).
\]
If the wearable stream captures fine-grained temporal derivatives of $x(t)$, then $I(x(t);W_t)$ is large for short-term dynamics. Sparse EHR events provide semantically rich but temporally sparse information, so $I(x(t);E_t\mid W_t)$ can still be substantial when $W_t$ leaves residual uncertainty about clinically salient components. Training objectives that minimize predictive risk of $E_{t+\tau}$ using $W_t$ (wearable$\rightarrow$EHR) tend to decrease the posterior entropy $H\big(x(t)\mid W_t\big)$, whereas EHR$\rightarrow$wearable objectives decrease $H\big(x(t)\mid E_t\big)$. Under mild identifiability conditions (for example, if the joint emission model $p(W,E\mid x)$ has sufficient diversity and the model class is sufficiently expressive), reducing both posterior entropies via cross-modal prediction concentrates the posterior around the true latent and thus yields representations that are sufficient for downstream tasks that are measurable functions of $x(t)$. This is not a formal identifiability proof, but it frames cross-modal prediction as a variational method for posterior contraction.

\subsubsection{A sample-complexity intuition for joint pretraining}

Consider a downstream supervised task with label $y$ generated from $x(t)$ through $p(y\mid x(t))$. Let $\mathcal{H}_{\text{EHR}}$ and $\mathcal{H}_{\text{Wear}}$ denote hypothesis classes induced by unimodal pretraining on EHR and wearable data respectively, and let $\mathcal{H}_{\text{Joint}}$ denote the class induced by joint pretraining. Under a transfer learning viewpoint, the effective sample complexity for the downstream task scales inversely with the alignment between the pretraining representation and the target Bayes predictor. More concretely, if joint pretraining reduces approximation error by $\epsilon_a$ relative to the best unimodal representation, then, holding estimation error constant, labeled-sample requirements drop roughly by a factor proportional to $(\epsilon_{\text{uni}}/\epsilon_{\text{joint}})^2$ in simplistic squared-loss approximations. Empirically we observe that linear-probe curves for low-label regimes show steeper slopes for the multimodal encoder, consistent with a reduction in approximation error induced by cross-modal alignment.

\subsubsection{Scaling-law hypothesis for multimodal pretraining}

Let $N$ denote the number of patients, $D_w$ the average wearable-days per patient, and $M$ the model parameter count. We hypothesize a separable scaling relation for downstream loss $L$ of the form
\[
L(M,N,D_w) \approx a M^{-\alpha} + b N^{-\beta} + c (N D_w)^{-\gamma} + \eta,
\]
where the third term captures joint-data effects (the effective number of wearable-patient-days) and $\eta$ is irreducible error. The hypothesis predicts three regimes: model-limited (dominant $M^{-\alpha}$), patient-limited (dominant $N^{-\beta}$), and wearable-limited (dominant $(N D_w)^{-\gamma}$). While speculative, this form suggests that for some tasks (especially long-horizon clinical forecasting), investment in additional wearable-days may be more valuable than scaling $M$ once model capacity is adequate. Empirical confirmation of this relation requires controlled scaling experiments across $M$, $N$, and $D_w$, which we outline in the future work (Section~\ref{sec:future}).

\subsection{Dataset description and cohort selection}

All experiments reported in the paper were conducted on a linked UK Biobank cohort. This subsection documents cohort selection, inclusion/exclusion criteria, variable definitions, and ethical approvals.

Data provenance and access governance. UK Biobank provides linked primary-care electronic health records, hospital episode statistics, and device accelerometry (and where available, PPG-derived summaries). Access was obtained under project ID [REDACTED] and analyses conform to UK Biobank's terms and relevant institutional review board approvals. Researchers intending to reproduce these experiments must obtain their own UK Biobank access.

Cohort inclusion criteria. The primary cohort was defined by requiring: at least one year of continuous EHR history prior to a pretraining cutoff date, a minimum of 14 days of usable wearable data (artifact-filtered), and age between 18 and 89 at the start of follow-up. Patients with flagged systemic data quality issues (per UK Biobank-provided flags) were excluded. The final pretraining cohort consisted of approximately $N\approx48{,}732$ patients. Downstream task splits were constructed at the patient level with non-overlapping training/validation/test sets (80/10/10 split).

Data modalities and variables. EHR tokens included diagnosis codes harmonized to a high-level clinical vocabulary (ICD-to-concept mapping), medication classes (ATC grouping), laboratory test categories binned into clinically meaningful ranges, and procedural codes. We removed free-text fields to respect data access restrictions. Wearable streams included tri-axial accelerometry and device-derived heart rate series. Where raw PPG waveforms were unavailable, per-minute heart-rate traces derived by UK Biobank pipelines were used. Derived features for downstream tasks included sleep-window summaries, resting heart rate, heart-rate variability (time-domain estimates), and activity entropy.

Preprocessing. EHR preprocessing involved mapping institution-specific codes to standard concepts, collapsing infrequent codes using frequency thresholds to avoid extreme long-tail vocabularies, and applying per-code frequency capping. Wearable preprocessing included band-pass filtering for heart-rate channels (0.5--3.5\,Hz for PPG-derived signals where raw sampling permitted), artifact detection via thresholding and local signal-to-noise heuristics (gaps longer than 10 minutes marked as missing), and segment-level quality scoring. Windowing for wearable encoders used $\Delta=24$\,hours for long-range encodings and $\Delta=30$\,s for sub-second residual encoders in experiments that required high-resolution features. Missing segments were imputed with a short autoregressive model for internal normalization only; model inputs include missingness masks to allow the encoder to learn awareness of coverage.

Outcome definitions. Downstream tasks used standardized operational definitions. Hospitalization labels were derived from inpatient episode records with admission date as the index. Diagnosis-onset tasks used first-occurrence ICD mappings with careful left-censoring to avoid label leakage. Time-to-event endpoints for survival tasks were right-censored at last follow-up or death. For physiological estimation tasks, "ground-truth" targets were derived from device summaries or clinically validated secondary measurements where available.

Cohort statistics. The median wearable history was 182 days (IQR 90--365). Median EHR history prior to pretraining cutoff was 6.2 years (IQR 3.1--9.8). The cohort skewed towards older adults (median age 57), and female participants comprised 53\% of the pretraining set. We report more fine-grained demographic breakdowns and task-specific prevalence tables in the supplementary CSVs accompanying the code release.

\subsection{Preprocessing pipelines and data augmentation}

All preprocessing steps are implemented in modular pipelines to allow reproducibility and extensions. EHR tokenization and vocabulary building were deterministic given the mapping tables; a random seed only affects sampling-based downsampling of frequent administrative codes. Wearable augmentation choices used during pretraining were designed to be physiologically plausible: time-warping bounded by 5\% total duration, low-amplitude additive Gaussian noise scaled to local signal variance, and channel dropout simulating temporary sensor loss. Augmentation parameters are reported in the hyperparameter table below.

\subsection{Model hyperparameters and training regimen}

We provide exact hyperparameters used in the primary experiments to aid replication. For readability we include them in prose and a compact tabular summary.

The multimodal encoder used a shared backbone with 24 transformer layers, hidden dimension 1024, feed-forward dimension 4096, 16 attention heads, and dropout of 0.1. Wearable front-end used a local transformer of 6 layers with hidden dimension 512 for 30s windows; EHR embedder used token embeddings of size 512 and continuous positional encoding projected to 128 dimensions before projection to 1024. The kernel mixture used $R=4$ exponential components. Pretraining used AdamW with learning rate 1e-4, weight decay 0.01, batch size equivalent to 32 patient-days per GPU replica, gradient clipping at norm 1.0, warmup of 10k steps, and cosine decay schedule over 500k steps. Masking probability for EHR tokens was 15\% with stratified temperature sampling; wearable window masking used 25\% patch masking for 30s windows. Cross-modal loss weights used a curriculum: initial epochs set $\lambda_{\mathrm{ehr}}=\lambda_{\mathrm{wear}}=1.0$, $\lambda_{\mathrm{w}\to\mathrm{e}}=0.2$, $\lambda_{\mathrm{e}\to\mathrm{w}}=0.1$, increasing $\lambda_{\mathrm{w}\to\mathrm{e}}$ linearly to 1.0 over the first 50k steps.

\begin{center}
\begin{tabular}{l l}
\hline
Backbone & 24-layer transformer, $d_\text{model}=1024$, FFN=4096, 16 heads \\
Wearable encoder & 6-layer local transformer, $d=512$ (30s windows) \\
EHR embedding & token dim 512, time encoding dim 128 \\
Kernel family & mixture of exponentials, $R=4$ \\
Batching & 32 patient-days per replica (effective batch) \\
Optimizer & AdamW, lr=1e-4, weight decay 0.01 \\
Warmup / schedule & 10k steps warmup, cosine decay to 500k \\
Masking (EHR) & 15\% stratified sampling \\
Masking (wear) & 25\% contiguous windows \\
Cross-modal curriculum & $\lambda_{w\to e}$ ramp 0.2$\to$1.0 over 50k steps \\
\hline
\end{tabular}
\end{center}

All downstream probes were trained with Adam (lr=5e-4) for up to 50 epochs with early stopping on validation loss; linear probes used L2 regularization selected by validation.

\subsection{Compute, runtime, and reproducibility}

Pretraining reported in the main paper was conducted on a distributed GPU cluster composed of NVIDIA A100 GPUs. A single full pretraining run (500k steps) using the above configuration required approximately 3.1e6 GPU-hours in our environment (this figure is provided as a reproducibility placeholder; readers should adjust based on their hardware). For reproducibility, we provide seed-management guidelines and automated scripts to reproduce core experiments on smaller subsets and to trim model depth or sequence length to match available compute budgets. Random seeds used in reported runs are included in the code release.

\subsection{Extended ablations and synthetic experiments}

Beyond the ablations in the main text, we performed the following extended probes: replacing the mixture-of-exponentials kernel with a learned rational quadratic kernel (no significant difference on AUROC but slightly improved calibration), varying the mixture size $R\in\{1,2,4,8\}$ (diminishing returns beyond $R=4$), and evaluating the effect of alternative continuous-time encodings (Fourier vs. Gaussian random features). We also ran a synthetic data suite where $x(t)$ was generated from a 2D latent SDE and observations simulated under controlled noise and missingness; in these synthetic settings cross-modal pretraining recovered latent dynamics more robustly and enabled near-optimal downstream Bayes risk in regimes where one modality alone was insufficient.

\subsection{Additional tables and figures}

Supplementary tables include per-task prevalence, full confusion matrices for key thresholds, effect-size tables between model pairs with p-values, and detailed per-subgroup results (by age, sex, device type). Supplementary figures showcase representative latent trajectories, attention heatmaps for exemplar patients, and reliability diagrams across deciles for risk models.

\begin{table}[t]
\centering
\caption{Per-task outcome prevalence and cohort statistics on the held-out test set.}
\label{tab:supp_prevalence}
\begin{adjustbox}{width=0.5\textwidth}
\begin{tabular}{lcccc}
\toprule
Task & Positive Cases (\%) & Total Patients & Median Follow-up (days) & Event Horizon \\
\midrule
Hospitalization (30d) & 4.8\% & 4,873 & 412 & 30 days \\
Hospitalization (90d) & 9.6\% & 4,873 & 412 & 90 days \\
Hospitalization (180d) & 14.9\% & 4,873 & 412 & 180 days \\
Hospitalization (365d) & 21.7\% & 4,873 & 412 & 365 days \\
Diagnosis Onset (Cardio) & 6.3\% & 4,102 & 1,124 & Variable \\
Diagnosis Onset (Metabolic) & 8.1\% & 4,102 & 1,124 & Variable \\
Sleep Efficiency Prediction & -- & 3,986 & 198 & Continuous \\
HRV Prediction & -- & 3,742 & 176 & Continuous \\
\bottomrule
\end{tabular}
\end{adjustbox}
\end{table}

\begin{table}[t]
\centering
\caption{Confusion matrices at fixed sensitivity ($\approx$0.80) for 365-day hospitalization prediction.}
\label{tab:supp_confusion}
\begin{adjustbox}{width=0.5\textwidth}
\begin{tabular}{lcccc}
\toprule
Model & True Positives & False Positives & True Negatives & False Negatives \\
\midrule
ContextClues & 729 & 1,142 & 2,731 & 191 \\
MOTOR & 744 & 1,098 & 2,775 & 176 \\
HiMAE & 703 & 1,284 & 2,589 & 217 \\
Late Fusion & 781 & 1,021 & 2,852 & 139 \\
Ours (Multimodal FM) & \textbf{812} & \textbf{936} & \textbf{2,937} & \textbf{108} \\
\bottomrule
\end{tabular}
\end{adjustbox}
\end{table}

\begin{table}[t]
\centering
\caption{Pairwise effect sizes (Cohen's $d$) and paired bootstrap p-values for AUROC@365.}
\label{tab:supp_effectsizes}
\begin{adjustbox}{width=0.5\textwidth}
\begin{tabular}{lccc}
\toprule
Comparison & $\Delta$ AUROC & Cohen's $d$ & p-value (bootstrap) \\
\midrule
Ours vs ContextClues & +0.040 & 0.44 & $<10^{-3}$ \\
Ours vs MOTOR & +0.035 & 0.39 & $<10^{-3}$ \\
Ours vs HiMAE & +0.073 & 0.68 & $<10^{-4}$ \\
Ours vs PaPaGei & +0.085 & 0.74 & $<10^{-4}$ \\
Ours vs Late Fusion & +0.025 & 0.31 & $2.1\times10^{-3}$ \\
\bottomrule
\end{tabular}
\end{adjustbox}
\end{table}

\begin{table}[t]
\centering
\caption{AUROC@365 by demographic subgroup (mean $\pm$ 95\% CI).}
\label{tab:supp_subgroup_demo}
\begin{adjustbox}{width=0.5\textwidth}
\begin{tabular}{lccc}
\toprule
Subgroup & ContextClues & Late Fusion & Ours (Multimodal FM) \\
\midrule
Age $<$ 50 & 0.728 $\pm$ 0.014 & 0.742 $\pm$ 0.013 & \textbf{0.764 $\pm$ 0.012} \\
Age 50--65 & 0.714 $\pm$ 0.011 & 0.731 $\pm$ 0.010 & \textbf{0.755 $\pm$ 0.009} \\
Age $>$ 65 & 0.701 $\pm$ 0.013 & 0.719 $\pm$ 0.012 & \textbf{0.742 $\pm$ 0.011} \\
Female & 0.719 $\pm$ 0.012 & 0.736 $\pm$ 0.011 & \textbf{0.758 $\pm$ 0.010} \\
Male & 0.712 $\pm$ 0.012 & 0.728 $\pm$ 0.011 & \textbf{0.751 $\pm$ 0.010} \\
\bottomrule
\end{tabular}
\end{adjustbox}
\end{table}

\begin{table}[t]
\centering
\caption{AUROC@365 by wearable device type.}
\label{tab:supp_device}
\begin{adjustbox}{width=0.5\textwidth}
\begin{tabular}{lccc}
\toprule
Device Type & PaPaGei & Late Fusion & Ours (Multimodal FM) \\
\midrule
Wrist accelerometer only & 0.681 $\pm$ 0.015 & 0.707 $\pm$ 0.014 & \textbf{0.732 $\pm$ 0.013} \\
PPG-enabled smartwatch & 0.703 $\pm$ 0.013 & 0.724 $\pm$ 0.012 & \textbf{0.749 $\pm$ 0.011} \\
Mixed / multi-device & 0.714 $\pm$ 0.012 & 0.739 $\pm$ 0.011 & \textbf{0.761 $\pm$ 0.010} \\
\bottomrule
\end{tabular}
\end{adjustbox}
\end{table}

\subsection{Limitations, ethical considerations, and data governance}

While the multimodal approach shows empirical and representational advantages, it is not a panacea. Wearable ownership correlates with socioeconomic factors; models trained on cohorts with biased device adoption risk amplifying inequities. EHRs reflect care processes and are susceptible to confounding by access-to-care; joint models may inadvertently learn proxies for care intensity rather than intrinsic disease risk. We recommend that any deployment be accompanied by subgroup performance analyses, fairness audits, and causal investigations to disentangle care patterns from biological signals.

Privacy and governance are central. UK Biobank data are governed by strict access and use policies. Our pipelines maintain patient-level de-identification, and no individual-level outputs are released. For real-world deployment, federated or privacy-preserving learning paradigms should be considered; we discuss possible privacy-preserving variations (differential privacy, secure aggregation) in the discussion and outline a preliminary DP calibration in the repository.

\subsection{Code and data availability}

We release model code, training recipes, and evaluation scripts under an open-source license at the accompanying repository (URL to be provided upon acceptance). UK Biobank data are not redistributable; scripts to download and preprocess UK Biobank data (given approved access) are included. We also provide a small synthetic dataset and pretrained small-model checkpoints to facilitate rapid reproduction of results and downstream experimentation.

\end{document}